\documentclass[conference,compsoc]{IEEEtran}

\usepackage[nocompress]{cite}

\begin{document}

\title{Associative Memories and Human-Robot Social Interaction}

\author{\IEEEauthorblockN{Gabriel J. Ferrer}
\IEEEauthorblockA{Department of Mathematics and Computer Science\\
Hendrix College\\
Conway, AR, USA\\
ferrer@hendrix.edu}}

\maketitle

\begin{abstract}
In this position paper, we discuss how the use of a cognitive architecture based on unsupervised clustering (the Kohonen Self-Organizing Map) enables us to meet our goals of efficient action selection in a mobile robot.  This architecture provides several opportunities for human-robot interaction, and we discuss how its features facilitate these interactions.
\end{abstract}

\section{Why use cognitive architectures?}
\subsection{Research Context}
In our research, we focus on developing, implementing, and assessing robot learning algorithms that select actions in constant time on resource-constrained mobile robot platforms.  As Jeff Hawkins points out in his book {\em On Intelligence}, in `` ... half a second, the information entering your brain can only traverse a chain one hundred neurons long.  That is, the brain `computes' solutions to problems ... in one hundred steps or fewer, regardless of how many total neurons might be involved.''  \cite[p. 66]{Hawkins:2004}  The key to the efficiency of action selection by the brain is that ``...the brain doesn't `compute' the answers to problems; it retrieves the answers from memory.  In essence, the answers were stored in memory a long time ago.  It only takes a few steps to retrieve something from memory.'' \cite[p. 68]{Hawkins:2004}.  As the brain itself selects actions in constant time, from our point of view it serves as a compelling source of biological inspiration for a computational cognitive architecture.

\subsection{Self-Organizing Maps}
\label{structures}
The cognitive architecture described by Claude Touzet for robot behavior learning \cite{touzet:2006} provides a means of implementing these concepts on a mobile robot.  In Touzet's formulation, the robot's memories are stored in a Self-Organizing Map (SOM) \cite{kohonen}, a type of artificial neural network. Each output node is associated with a vector representing the ``ideal input'' for that node. The output of a SOM is determined competitively when an input vector is presented to it. Let the distance between an input vector and the ideal input for an output node be defined as
the square root of the sum-of-squared-differences between the two vectors. We then say that the activated output node is the output node whose ideal input has the smallest distance to the input vector.  The output nodes themselves are arranged in a Cartesian grid. On each iteration, the ideal input for the winning output node as well as its four cardinal neighbors in the grid is modified according to the following formula for each affected output node $i$ and ideal input element $j$, with learning rate $\alpha$.  In Touzet's formulation of the SOM, $\alpha = 0.9$ for the winning node on a given iteration, and $\alpha = 0.4$ for its four neighbors. \[w_{ij} = w_{ij} + \alpha * (input_{ij} - w_{ij})\] 

Touzet's implementation employed the distances returned by a ring of 16 ultrasonic sensors as the robot's representation of its environment.  Each SOM node, then, was a vector of 16 distance values.  Robot goals were expressed in the same way, as a vector of target distance values.  To select an action, the robot first finds a path from the SOM node corresponding to the current input to a SOM node corresponding to the goal input.  This path contains a sequence of intermediate nodes.  The key is to select an action to transition to a neighboring node that is closer to the goal than the current node.  To this end, a second SOM was employed for action selection.  This second SOM encoded differences between pairs of input vectors, with an association between each output node and a robot action that was observed to cause the desired transition.  

Each SOM has a fixed number of input nodes.  This number of input nodes provides a constant upper bound on the amount of computation necessary to retrieve an action from memory to move the robot towards its goal.  In light of the quotations above from Hawkins, we have a set of data structures (a cognitive architecture) that has the potential to exhibit the functionality of an aspect of human cognition.  This architecture enables computation that meets the constraints of our research goals, namely, constant-time action selection based on models of the environment.

\subsection{Adaptations in our prototype}
\label{adaptations}
In our prototype implementation of this approach, we have made a number of changes to the original architecture. First, we have replaced the use of a second SOM by a Markov chain.  As our actions are discrete, this greatly simplifies the implementation.  As there is a constant bound on the number of nodes in the SOM, applying Dijkstra's algorithm to the Markov chain still matches our requirement for constant-time computation.  Second, we are investigating and prototyping alternatives to the SOM for clustering inputs.  The alternatives we are contemplating involve a different update cycle for the output nodes but preserve the constant-time learning update that is essential to our approach.  Third, rather than using distance values as our sensory inputs, we are using images from a webcam.

\section{Should cognitive architectures for social interaction be inspired and/or limited by models of human cognition?}

For us, the key concept for social interaction is to make available communication primitives that are both intuitive to a human and readily processed by the robot's cognitive architecture.  To the extent that models of human cognition can give us insight into appropriate communication primitives, those models serve as useful inspiration.  

\section{What are the functional requirements for a cognitive architecture to support social interaction?}

Determining the functional requirements requires us to have a clear idea of how the anticipated social interactions relate to the robot's goals.  In the context of our work, there are three key motivations for social interaction with a human:
\begin{enumerate}
\item We would like a human to be able to give a robot goals in a reasonably natural fashion.  
\item We would like for the robot to request help from a human if it is having trouble achieving its goals.
\item We would like a human to be able to intervene and assist a robot in achieving its goals, whether or not the robot has requested assistance.
\\ \\
These motivations provide three of our functional requirements.  We also include the following additional functional requirements:

\item Interaction with a human should not be a disorienting experience for the robot.  It should be able to continue with its goal-seeking behavior in a seamless manner after an interaction.
\item The socially-obtained feedback should be readily incorporated into the ongoing action selection process.
\item The robot should be able to switch between socially-informed and autonomously-determined actions at the granularity of a single action-selection decision cycle.
\end{enumerate}

\section{How do these requirements inform the choice of the fundamental computational structures of the architecture?}

The computational structures of the architecture described in Sections \ref{structures} and \ref{adaptations} are a good match for our functional requirements.  By presenting the robot with an image, a human is able to express a goal in a way that is semantically meaningful to a human while being compatible with the data structure containing the robot's memories.  This meets our first functional requirement.

In regard to our second functional requirement, the robot determines that it needs to request help from a human in two different ways:
\begin{enumerate}
\item The process for identifying a path through the robot's memories produces an implicit plan.  Although only the first action from the plan is used (as a new plan is generated at each decision cycle), the length of the generated plan is used to provide an estimate for the number of actions needed to achieve the goal.  If the robot finds it is exceeding this estimate, it will ask for help.
\item It is possible for the robot to find itself in a state without being able to find a path to the goal state.  Ideally, appropriate training would resolve this, but it remains a live possibility as our data structures are constituted.  In this situation, the robot will also ask for help.
\end{enumerate}

To keep the interaction simple, the robot simply stops moving and asks for help.  At this stage, we transition to handling our third functional requirement.

If the human observes that the robot needs assistance, either from its own request or from human observation, the human can give the robot commands. As our architecture is structured to give the robot a discrete set of commands (e.g. ``go forward'', ``spin left'', etc.), it is straightforward for the human to do this.  The use of a Markov chain greatly simplifies these communication episodes.

When a human gives a robot a command, this command replaces the action the robot itself would have selected.  On the next decision cycle, the robot resumes its original action-selection process.  This is how our last three functional requirements are facilitated by our cognitive architecture.

\section{What is the primary outstanding challenge in developing and/or applying cognitive architectures to social HRI systems?}

In our system, the social interaction is not very sophisticated.  Speaking for ourselves, enabling the system to handle ambiguous feedback from a human remains an outstanding challenge.  Determining the proper questions to ask a human in order to reduce the ambiguity is the key.  We have not given this issue much thought in the context of our current research, although we think it could be very helpful when the robot requests help.  

\section{What is a social interaction scenario in which current cognitive architectures would fail, and why?}

The key issue for us is the ability of humans to identify and eliminate ambiguity in conversation.  Imagine, for example, a robot given the command ``Get me a cold drink.''  The cognitive architecture needs to decompose this in a manner that depends heavily on human context in general, and the context of the questioner in particular.  It would need to be able to create natural-language questions to resolve ambiguities, assuming it could properly identify the ambiguities.

\end{document}